\documentclass[11pt,reqno]{amsart}

\usepackage{amsmath,amsthm,amssymb,enumitem,xspace}

\newcommand{\Z}{\mathbb{Z}}

\newcommand{\R}{\mathbb{R}}

\DeclareMathOperator{\conv}{conv}

\newcommand{\norm}[2]{\left\|#2\right\|_{#1}}

\DeclareMathOperator*{\trace}{Tr}

\newtheorem{proposition}{Proposition}
\newtheorem{theorem}{Theorem}

\newtheorem{lemma}{Lemma}
\theoremstyle{definition}

\theoremstyle{remark}
\newtheorem{remark}{Remark}

\usepackage{algorithm,algorithmicx}
\usepackage{algpseudocode}
\usepackage{bbm}
\usepackage{bm}
\usepackage[mathlines,pagewise,left]{lineno}
\usepackage{amssymb}
\usepackage{color,xcolor,tikz}
\usetikzlibrary{arrows,shapes,chains}  
\usepackage{amsmath}
\usepackage{bcol}
\usepackage{multirow, array}
\usepackage{xspace}
\usepackage[numbers]{natbib}
\usepackage{comment}
\usepackage[toc,page]{appendix}
\usepackage{graphicx} 
\usepackage{booktabs} 
\usepackage{pdflscape}
\usepackage[export]{adjustbox}
\usepackage{caption}
\usepackage[font=scriptsize]{subcaption}
\usepackage{mathtools}
\usepackage{hyperref}
\usepackage{amsthm}
\theoremstyle{definition}

\theoremstyle{remark}

\renewcommand{\conv}{\ensuremath{\text{cl conv}}}

\usepackage{stackengine,scalerel}
\newcommand{\dddagger}{%
	\text{$\sbox0{\ddag}\stretchrel*{%
		\stackengine{-.6\ht0}{\ddag}{\ddag}{O}{c}{F}{F}{S}}{\ddag}$%
}}

\newcolumntype{\resetRow}{>{\global\let\currentrowstyle\relax}}
\newcolumntype{^}{>{\currentrowstyle}}

\def\SingleSpacedXI{\linespread{1.1}}
\SingleSpacedXI

\usepackage[colorinlistoftodos,prependcaption,textsize=small]{todonotes}

\title[Robust SVMs via conic optimization]{Robust support vector machines via conic optimization}
\author{Valentina Cepeda$^\dagger$, Andr\'es G\'omez$^\ddagger$, Shaoning Han$^\dddagger$
  }
\thanks{ \noindent \hskip -5mm
	\hspace{-0.2em}$^\dagger$Department of Industrial Engineering, Universidad de los Andes, Bogot\'a, 
	 Colombia. \texttt{v.cepeda@uniandes.edu.co}\\
	 $^\ddagger$Daniel J. Epstein Department of Industrial and Systems Engineering, University of Southern California, Los Angeles, CA, USA. \texttt{gomezand@usc.edu}\\
	 $^\dddagger$Daniel J. Epstein Department of Industrial and Systems Engineering, University of Southern California, Los Angeles, CA, USA. \texttt{shaoning@usc.edu}
}
\begin{document}
\maketitle
\begin{center}
	February 2024
\end{center}
\begin{abstract}
	\vskip 3mm
	\noindent We consider the problem of learning support vector machines robust to uncertainty. It has been established in the literature that typical loss functions, including the hinge loss, are sensible to data perturbations and outliers, thus performing poorly in the setting considered. In contrast, using the 0-1 loss or a suitable non-convex approximation results in robust estimators, at the expense of large computational costs. In this paper we use mixed-integer optimization techniques to derive a new loss function that better approximates the 0-1 loss compared with existing alternatives, while preserving the convexity of the learning problem. In our computational results, we show that the proposed estimator is competitive with the standard SVMs with the hinge loss in outlier-free regimes and better in the presence of outliers.
	
	\noindent
	\textbf{Keywords}. Support vector machine, robustness, mixed-integer nonlinear optimization, convexification, indicator variables. \\
\end{abstract}

\section{Introduction}

Given labeled data $\{(\bm{x_i},y_i)\}_{i=1}^n$, where $\bm{x_i}\in \R^p$ encodes the features of point $i$ and $y_i\in \{-1,1\}$ denotes its class, consider the support vector machine (SVM) problem defined in the seminal paper by 
\citet{cortes1995support}, given by
\begin{subequations}\label{eq:SVM}
	\begin{align}
		\min_{\bm{w}\in \R^p,\bm{\xi}\in \R_+^n}\;&\|\bm{w}\|_2^2+\lambda\sum_{i=1}^n \xi_i^\eta\\
		\text{s.t.}\:&y_i\left(\bm{x_i^\top w}\right)\geq 1-\xi_i\quad \forall i\in \{1,\dots,n\}
	\end{align}
	
\end{subequations}
\noindent for some regularization parameter $\lambda>0$ and some sufficiently small constant $\eta\in \R_+$. \citet{cortes1995support} argue that if $\lambda$ is sufficiently large and $\eta$ is sufficiently small, then an optimal solution of \eqref{eq:SVM} yields a hyperplane that misclassifies the least amount of points possible and has maximum margin among all hyperplanes with minimal misclassification. However, noting that \eqref{eq:SVM} is NP-hard if $\eta<1$, the authors instead suggest using $\eta=1$, resulting in the popular SVM with the hinge loss
\begin{align}\label{eq:hinge}
	\min_{\bm{w}\in \R^p,\bm{\xi}\in \R_+^n}\;&\|\bm{w}\|_2^2+\lambda\sum_{i=1}^n \max\left\{0,1-y_i\left(\bm{x_i^\top w}\right)\right\}.
\end{align}
Typical convex surrogates of the 0-1 or misclassification loss --corresponding to setting $\eta=0$ in \eqref{eq:SVM}-- have been shown to perform well in low noise settings \citep{bartlett2006convexity}. However, these surrogates are known to perform worse than the 0-1 loss in the presence of uncertainty or outliers \citep{manwani2013noise,ghosh2015making}.

\begin{figure*}[!t]
	\centering
\subfloat[ Normalized sigmoid ]{\includegraphics[width=0.26\textwidth,trim={10cm 5cm 11cm 5cm},clip]{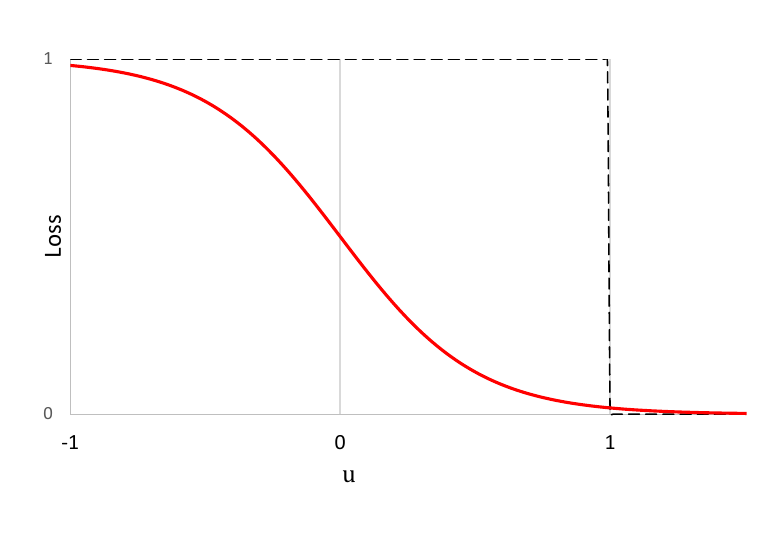}}\hfill
	\subfloat[$\psi$-learning]{\includegraphics[width=0.26\textwidth,trim={10cm 5cm 11cm 5cm},clip]{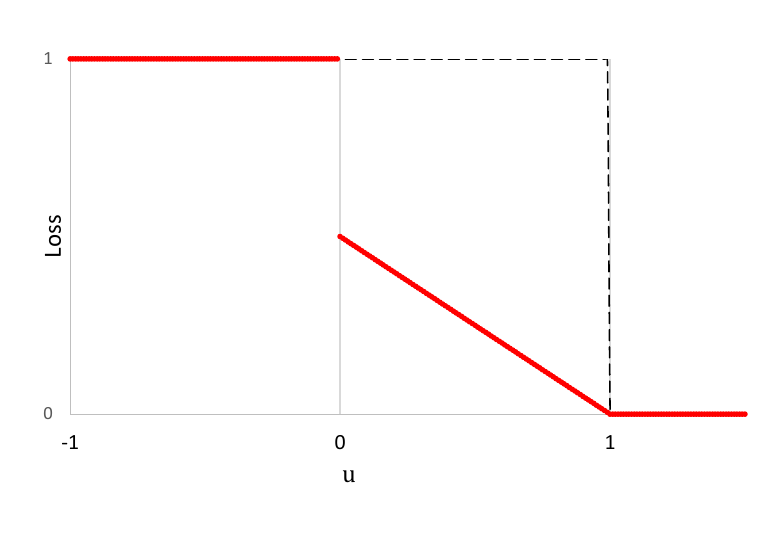}}\hfill
	\subfloat[ Ramp ]{\includegraphics[width=0.26\textwidth,trim={10cm 5cm 11cm 5cm},clip]{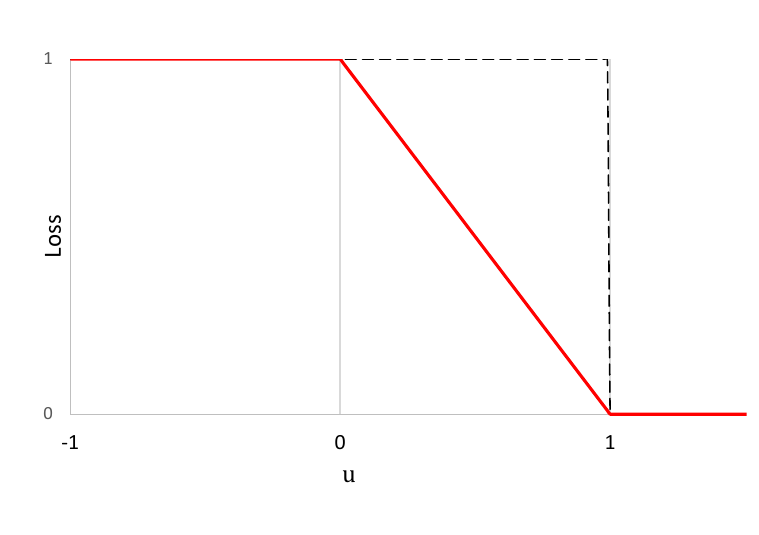}}
	\newline
	\subfloat[ Conic with $\gamma_i=0.2$ ]{\includegraphics[width=0.26\textwidth,trim={10cm 5cm 11cm 5cm},clip]{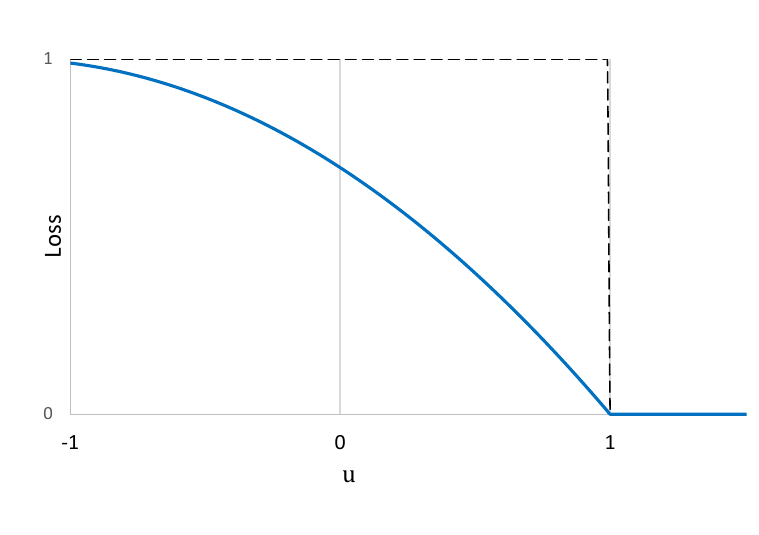}}\hfill
	\subfloat[ Conic with $\gamma_i=0.5$ ]{\includegraphics[width=0.26\textwidth,trim={10cm 5cm 11cm 5cm},clip]{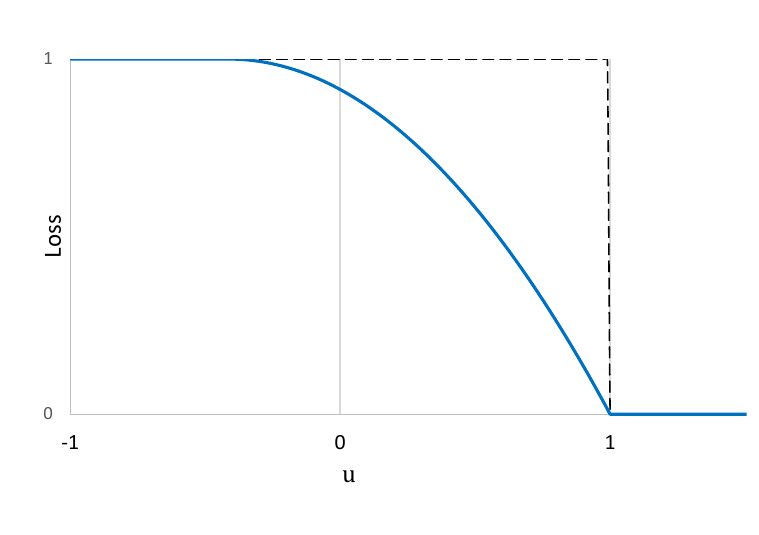}}\hfill
	\subfloat[ Conic with $\gamma_i=1.0$ ]{\includegraphics[width=0.26\textwidth,trim={10cm 5cm 11cm 5cm},clip]{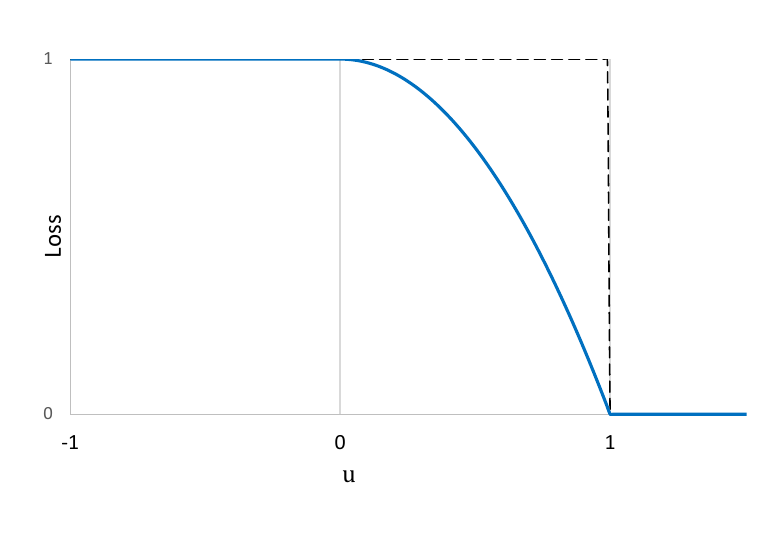}}\hfill
	\caption{\small Non-convex robust losses approximating the 0-1 loss, as a function of $u=y_i\bm{x_i^\top w}$. Top row: loss functions from the literature: the normalized sigmoid loss \citep{mason1999boosting}, the $\psi$-learning loss \citep{shen2003psi} and the ramp loss \citep{wu2007robust}. Bottom row: the proposed conic loss in Proposition~\ref{prop:loss} for different values of hyperparameters $\gamma_i$ (with $\lambda=1$) . By solving the conic optimization problem \eqref{eq:primal}, $\bm{\gamma}$ is chosen automatically to ensure convexity of the ensuing learning problem.}
	\label{fig:lossFunctions}
\end{figure*}

In particular, label noise, characterized by label flips in data, affects the performance of classification problems. The standard hinge loss SVM algorithm \eqref{eq:hinge} tends to penalize outliers more and, as a result, these points strongly influence the placement of the separating hyperplane, leading to low classification performance. To improve SVM robustness to label uncertainty, various approaches have been proposed in the literature. \citet{song2002robust} suggested evaluating an adaptive margin based on the distance of each point to its class center to mitigate the influence of points far from class centroids. 
Other studies have focused on replacing the hinge loss function with non-convex alternatives. Some popular choices are the ramp loss \citep{wu2007robust}, the rescaled hinge loss \citep{xu2017robust}, the $\psi$-learning loss \citep{shen2003psi} or simply using the exact 0-1 loss. Figure~\ref{fig:lossFunctions} (top row) depicts these loss functions, which can be interpreted as non-convex approximations of the 0-1 loss. Unfortunately, while improving robustness, these methods may result in prohibitive increases in computational time due to the non-convex nature of the optimization problems. For example, \citet{brooks2011support} propose to use mixed-integer optimization (MIO) solvers to tackle SVMs with the ramp or 0-1 loss using big-M formulations: the resulting approach, which we review in \S\ref{sec:bigM}, does not scale beyond $n\approx 100$. Thus, heuristics or local minimization methods which do not produce global minimizers are typically used in practice.

In this paper, we derive new non-convex loss functions for robust SVMs. The loss functions, depicted in Figure~\ref{fig:lossFunctions} (bottom), are better approximations of the 0-1 loss function than existing alternatives in the literature, and we also propose a method to tune the parameters of the loss function to ensure that
the SVM problem can be solved to optimality via conic optimization. The proposed approach is inspired by convexification techniques from the MIO literature. 

\subsection*{MIO, machine learning and convexification}
Several non-convex learning problems can be naturally cast as MIO problems \citep{carrizosa2013supervised}. In the context of SVMs, \citet{brooks2011support} proposes a MIO formulation to tackle problems with either the 0-1 or the ramp loss, using big-M reformulations to encode the non-convexities; we review this method in \S\ref{sec:bigM}. Other MIO approaches for SVMs include problems with the hinge loss but sparsity \citep{guan2009mixed,maldonado2014feature}, as well as the approach by \citet{ustun2013supersparse} which imposes sparsity and uses the 0-1 loss.

However, simply resorting to off-the-shelf MIO solvers may result in subpar performance. Indeed, these solvers are \emph{general-purpose}, able to solve a large variety of problems but not particularly tailored to any, and certainly not tailored to solve machine learning problems. Nonetheless, in selected learning problems, and in particular in the \emph{best subset selection problem} \cite{miller2002subset}, MIO technology can tackle problems with tens of thousands of decision variables \citep{atamturk2020safe,hazimeh2022sparse}. The key ingredient beyond these algorithms is the exploit of the perspective relaxation \citep{Gunluk2010}, that is, a strong convex relaxation of the non-convex learning problem that does not depend on Big-M terms. Interestingly, simply solving the perspective relaxation without the additional enumeration used by MIO methods (possibly with some rounding) results in an estimator in its own right \cite{pilanci2015sparse,xie2020scalable,bertsimas2020sparse,dong2015regularization}, closely related to the MCP estimator in the literature \cite{zhang2010nearly}. More sophisticated convex relaxations of MIO formulations have been successfully deployed in several machine learning problems including regression \cite{atamturk2019rank,gomez2021outlier}, matrix decomposition and completion \cite{bertsimas2023sparse,bertsimas2022mixed}, sparse principal component analysis \cite{d2004direct,dey2023solving,kim2022convexification,li2020exact} and K-means clustering \cite{de2022ratio,peng2005new}.

\subsection*{Contributions and outline} In this paper we use MIO technology to derive strong convex relaxations of SVMs with the 0-1 loss. The resulting formulation can be interpreted as a new non-convex loss function that: \textit{(i)} is separable; \textit{(ii)} underestimates the 0-1 loss; \textit{(iii)} preserves convexity of the SVM problem; and \textit{(iv)} is stronger than alternative loss functions proposed in literature with properties \textit{(i)}-\textit{(iii)}. Moreover, we show that SVM problems with the derived loss can be implemented as semidefinite programs (SDPs). 

The rest of the paper is organized as follows. In \S\ref{sec:bigM} we review existing MIO formulations and their connections with the hinge loss. In \S\ref{sec:convexHull} we derive the proposed loss function, in \S\ref{sec:primal} we discuss its implementation and in \S\ref{sec:computations} we present computational experiments.

\subsection*{Notation} In the paper, we denote vectors and matrices in $\textbf{bold}$. We let $\bm{0}$ and $\bm{1}$ be vectors of zeros and ones, respectively, whose dimensions can be inferred from the context. Given a number $k\in \Z_+$, define $[k]:=\{1,\dots,k\}$. Given $\alpha\in \R$, we let $(\alpha)_+:=\max\{\alpha,0\}$ and $(\alpha)_-:=\max\{-\alpha,0\}$ denote the positive and negative parts of $\alpha$, respectively, so that the identity $\alpha=(\alpha)_+-(\alpha)_-$ holds. Moreover, we use $(\alpha)_{\pm}^2$ as a shorthand for $\left((\alpha)_{\pm}\right)^2$, i.e., the positive/negative part operator is applied before squaring the term. For any set $S$, we denote the convex hull of $S$ by $\conv(S)$. For any $a\in\R$, we adopt the following convention of division by 0: $a^2/0=0$ if $a=0$ and $a^2/0=+\infty$ otherwise. For any two matrices $\bm A$ and $\bm B$, we use $\bm A\succeq \bm B$ to represent that $\bm A-\bm B$ is positive semidefinite, and denote by $\langle \bm{A},\bm{B}\rangle:=\sum_{i}\sum_{j}A_{ij}B_{ij}$. Given any logical condition ``$\cdot$", we use $\mathbbm{1}_{\{\cdot\}}$ to denote the function that is equal to $1$ whenever the condition is satisfied, and $0$ otherwise.

\section{The big-M formulation and hinge loss} \label{sec:bigM}
To solve SVMs with the exact misclassification loss, \citet{brooks2011support} proposes to introduce binary variables $z_i=1$ if point $i$ lies in the margin or is misclassified, and $z_i=0$ if the point is correctly 
classified. Then, they reformulate \eqref{eq:SVM} with $\eta=0$ as
\begin{subequations}\label{eq:BigMPenalty}
	\begin{align}
		\min_{\bm{w},\,\bm z}\;&\|\bm{w}\|_2^2+\lambda\sum_{i=1}^n z_i\label{eq:BigMPenalty_obj}\\
		\text{s.t.}\:&y_i\left(\bm{x_i^\top w}\right)\geq 1-Mz_i\quad \forall i\in [n],\label{eq:BigMPenalty_classification}\\
		&\bm{w}\in \R^p,\;\bm{z}\in \{0,1\}^n
	\end{align}
\end{subequations}
where $M$ is a sufficiently large number. If $z_i=0$, constraint \eqref{eq:BigMPenalty_classification} forces point $i$ to be correctly classified, while setting $z_i=1$ incurs a fixed cost of $\lambda$ but allows point $i$ to be in the margin or in the ``incorrect" side of the hyperplane. \citet{brooks2011support} also shows that the estimator obtained from solving \eqref{eq:BigMPenalty} to optimality is consistent under appropriate conditions. 

A good proxy for the effectiveness of branch-and-bound methods is the relaxation quality of the MIO formulation, obtained by relaxing the binary constraints to bound constraints $\bm{0}\leq \bm{z}\leq \bm{1}$. 

\subsection*{Relaxation quality} Consider the relaxation of \eqref{eq:BigMPenalty}:
\begin{subequations}\label{eq:BigMPenaltyRelaxation}
	\begin{align}
		\min_{\bm{w}\in \R^p,\bm{0}\leq\bm{z}\leq \bm{1}}\;&\|\bm{w}\|_2^2+\lambda\sum_{i=1}^n z_i\\
		\text{s.t.}\:&y_i\left(\bm{x_i^\top w}\right)\geq 1-Mz_i\quad \forall i\in [n].
	\end{align}
\end{subequations}\normalsize
A common approach to assess the relaxation quality is through the $\texttt{gap}:=\left(\zeta_{mio}-\zeta_{relax}\right)/\zeta_{mio}$, where $\zeta_{mio}$ and $\zeta_{relax}$ denote the optimal objective values of \eqref{eq:BigMPenalty} and \eqref{eq:BigMPenaltyRelaxation}, respectively. Unfortunately, \emph{regardless of the data}, \eqref{eq:BigMPenaltyRelaxation} results in almost trivial relaxation gaps. Indeed, note that the solution $\bm{w}=\bm{0}$, $\bm{z}=1/M$ is always feasible with an objective value of $(\lambda n)/M\xrightarrow{M\to\infty}0$. Thus, if a large value of $M$ is used, the gap is close to $1$. Additionally, the solutions $(\bm{w},\bm{z})$ obtained from solving the relaxation are uninformative, which will severely impair branch-and-bound algorithms.  \citet{brooks2011support} reports that in the pool of 1,425 instances tested in their paper, while 39\% are solved to optimality, the average gap on the remaining 61\% instances is over 70\% after 10 minutes of branch-and-bound. We performed computations with this formulation using the commercial solver Gurobi, and observed a similar lackluster performance: instances with $n=100$ and $p=3$ cannot be solved in 10 minutes, and no instance with $n=200$ could be solved (typically resulting in gaps above 50\%).

\subsection*{Connections to the hinge loss} The relaxation \eqref{eq:BigMPenaltyRelaxation} is connected with problem \eqref{eq:hinge}. Indeed, note that in optimal solutions of the relaxation \eqref{eq:BigMPenaltyRelaxation} we find that $M z_i^*=\left(1-y_i\left(\bm{x_i^\top w}\right)\right)_+$, and thus (if $M$ is big enough), problem \eqref{eq:BigMPenaltyRelaxation} is equivalent to \begin{align*}
	\min_{\bm{w}\in \R^p,\bm{\xi}\in \R_+^n}\;&\|\bm{w}\|_2^2+\frac{\lambda}{M}\sum_{i=1}^n \max\left\{0,1-y_i\left(\bm{x_i^\top w}\right)\right\}.
\end{align*}
Thus, we see that SVMs with the hinge loss \eqref{eq:hinge} can be interpreted as an approximation\footnote{Problem \eqref{eq:hinge} is only a relaxation if the regularization controlling the hinge loss is negligible. Since practical uses of \eqref{eq:BigMPenalty} use much larger parameters, i.e., values of $M$ too small to guarantee the correctness of the MIO formulation, we prefer the term ``approximation" instead.} of \eqref{eq:BigMPenalty} informed by its natural continuous relaxation.

\subsection*{Alternative non-convex reformulations} Observe that we can rewrite the misclassification constraints for point $i\in [n]$ as the quadratic non-convex constraints
\begin{subequations} \label{eq:nonlinearClassification}
	\begin{align}
		(1-y_i\left(\bm{x_i^\top w}\right))(1-z_i)&\leq 0\label{eq:nonlinearClassification_1}\\
		(1-y_i\left(\bm{x_i^\top w}\right))z_i&\geq 0\label{eq:nonlinearClassification_2}.
	\end{align}
\end{subequations}
Constraints \eqref{eq:nonlinearClassification_1} correspond exactly to \eqref{eq:BigMPenalty_classification}; constraints \eqref{eq:nonlinearClassification_2}, which force misclassified to be in the ``incorrect" side of the hyperplane, are valid (but redundant) for the core SVM problem, and we keep them in the formulation for generality. While reformulation~\eqref{eq:nonlinearClassification} does not offer an advantage in terms of MIO methods, we use these constraints in the rest of the paper as they exactly represent the non-convexities of SVM problems without introducing artificial bounds $M$.

\section{Loss functions via convexification}\label{sec:convexHull}
Consider the problem of designing loss functions $\mathcal{L}_i:\R\to\R_+$ (which may depend on the point $i\in [n]$) such that 
\begin{equation}\label{eq:separable}\nu^*=\min_{\bm{w}\in \R^p}\;\|\bm{w}\|_2^2+\sum_{i=1}^n \mathcal{L}_i(y_i\bm{x_i^\top w})
\end{equation}
is a \emph{strong} (i.e., the objective $\nu^*$ should be as large as possible) and \emph{convex} relaxation of the SVM problem with 0-1 loss (i.e., $\mathcal{L}_i(y_i\bm{x_i^\top w})\leq \lambda \mathbbm{1}_{\{y_i\bm{x_i^\top w}<1\}}$). Note that these two goals are often at odds with each other. A natural approach to ensure convexity is forcing all loss functions to be convex themselves: however, in that case (and assuming the existence of an upper bound on the maximum margin $y_i\bm{x_i^\top w}-1$) the best possible relaxation is the hinge/big-M relaxation which, as discussed in \S\ref{sec:bigM}, is weak -- and reduces to the trivial lower bound $\mathcal{L}_i(\cdot)=0$ as the upper bound goes to infinity. Alternatively, the non-convex losses depicted in Figure~\ref{fig:lossFunctions} (except the sigmoid loss, which is not an underestimator) are stronger, but in general result in non-convex problems. The key idea is thus to carefully design loss functions that are non-convex, but the induced non-convexities are offset by the strong convexity of the term $\|\bm{w}\|_2^2$. 

Consider the extreme case where function $\mathcal{L}_i$ subsumes all non-convexities of the problem and thus $\mathcal{L}_j=0$ for all $j\neq i$. In this case, the best loss function $\mathcal{L}_i$ can be obtained from the convex envelope of the function $\psi(\bm{w}):=\|\bm{w}\|_2^2+\lambda\mathbbm{1}_{\{y_i\bm{x_i^\top w}<1\}}$.
In this section, we derive this convex envelope by studying the more general mixed-integer set 
\begin{equation*}  
		\begin{aligned}W_Q=&\left\{\bm{w}\in \R^p,\;z\in \{0,1\},\;t\in \R:
			t\geq \bm{w^\top Q w},\vphantom{\Big(1-y(\bm{x^\top w})\Big)}\right.\\
			&\qquad\left.\Big(1-y(\bm{x^\top w})\Big)(1-z)\leq 0,\;\Big(1-y(\bm{x^\top w})\Big)z\geq0\right\},
		\end{aligned}
\end{equation*}
where $\bm{Q}\succeq 0$ and where we dropped the index $i$ from $W_Q$ for simplicity. 
Indeed, observe that the epigraph of $\psi$ can be written in terms of $W_Q$ by setting $\bm Q=\bm I$, since 
\begin{align*}
	\left\{(\bm{w},\tau):\psi(\bm{w})\leq \tau\right\}
	\Leftrightarrow \left\{(\bm{w},\tau):\exists (z,t) \text{ s.t. }t+\lambda z\leq \tau,\; (\bm{w},z,t)\in W_I\right\}.
\end{align*}
The rest of the section is devoted to studying $\conv(W_Q)$.

\subsection{Derivation of the convex hull}

This subsection is devoted to proving Theorem~\ref{theo:convexHull} below.
\begin{theorem}\label{theo:convexHull}
	The convex hull of $W_Q$ is described by bound constraints $0\leq z\leq 1$, and inequality
	\begin{align*} t\geq \bm{w^\top Qw}-\gamma\Big(1-y(\bm{x^\top w})\Big)^2+\gamma\frac{\Big(1-y(\bm{x^\top w})\Big)_+^2}{z}+\gamma\frac{\Big(1-y(\bm{x^\top w})\Big)_-^2}{1-z},
	\end{align*}
	where $\gamma\in \R_+$ is the largest number such that $\bm{w^\top Qw}-\gamma\Big(1-y(\bm{x^\top w})\Big)^2$ is convex, given by $\gamma=1/(\bm{x^\top Q^{-1}x})$.
\end{theorem}

The rest of this subsection can be safely skipped by readers not interested in the proof of the main theorem.
We first establish the convexification result for a simplified version of $W_Q$ with only three variables. Given $b\in\R$, define 
\begin{align*}\widehat W=\left\{(w,z,t):\;
	t\geq w^2,\;z\in\{0,1\},\;  (w-b)(1-z)\ge0,\;(w-b)z\le 0 \right\}.
\end{align*}
\begin{lemma}\label{lem:cvxHull}
	The convex hull of $\widehat W$ described by $0\le z\le 1$ and 
	\begin{align}\label{eq:lemCvxHull}
		t\ge \frac{(w-b)_+^2}{1-z}+\frac{(w-b)_-^2}{z}+2bw-b^2.
	\end{align}
\end{lemma}
\begin{proof}
	By definition, $(w,z,t)\in\conv(\widehat W)$ if and only if there exists $0\le \lambda\le 1$ and $(t_0,w_0,0),$ $(t_1,w_1,1)\in \widehat W$, such that $(t,w,z)=(1-\lambda)(t_0,w_0,0)+\lambda (t_1,w_1,1)$. Equivalently, we find that $z=\lambda$ and the following inequality system is consistent
	\begin{align*}
		\begin{aligned}
			&t=(1-z)t_0+zt_1,\;w=(1-z)w_0+zw_1,\\
			&t_0\ge w_0^2,\;w_0-b\ge0,\\
			&t_1\ge w_1^2,\;w_1-b\le 0.     
		\end{aligned}
	\end{align*}
	By substituting out $t_i$ with their lower bounds $w_i^2$, $i=1,2$, we find the equivalent system    \begin{align}\label{eq:extendedLemCvxHull}
		\begin{aligned}
			&t\ge(1-z)w_0^2+zw_1^2,\;w=(1-z)w_0+zw_1,\\
			&\;w_0-b\ge0,\;w_1-b\le 0.       
		\end{aligned}
	\end{align}  
	It suffices to prove by cases that after eliminating additional variables $(w_0,w_1)$, \eqref{eq:extendedLemCvxHull} is equivalent to \eqref{eq:lemCvxHull}.
	\begin{itemize}[leftmargin=*]
		\item \emph{Case 1: $z=0$.} In this case, by the convention of division by 0, \eqref{eq:lemCvxHull} reduces to $t\ge (w-b)_+^2+2bw-b^2$ and $(w-b)_-=0$, which is equivalent to $w\ge b$ and $t\ge (w-b)_+^2+2bw-b^2=w^2$, or namely, \eqref{eq:extendedLemCvxHull}. 
		\item \emph{Case 2: $z=1$.} In this case, by the convention of division by 0, \eqref{eq:lemCvxHull} reduces to $t\ge (w-b)_-^2+2bw-b^2$ and $(w-b)_+=0$, which is equivalent to $w\le b$ and $t\ge (w-b)_-^2+2bw-b^2=w^2$, or namely, \eqref{eq:extendedLemCvxHull}. 
		\item \emph{Case 3: $0<z<1$.} Since $w=(1-z)w_0+zw_1=(1-z)(w_0-b)+z(w_1-b)+b$, one has
		\begin{align*}
			&(1-z)w_0^2+zw_1^2\\
			=&(1-z)(w_0-b)^2+z(w_1-b)^2+b^2+2b[(1-z)(w_0-b)+z(w_1-b)]\\
			=&(1-z)(w_0-b)^2+z(w_1-b)^2+b^2+2b(w-b)\\
			=&(1-z)(w_0-b)^2+z(w_1-b)^2+2bw-b^2.
		\end{align*} By changing variables $p=(w_0-b)(1-z)$ and $q=-(w_1-b)z$, eliminating $(w_0,w_1)$ in \eqref{eq:extendedLemCvxHull} amounts to
		\begin{align*}
			t\ge \min_{p,q}\;&\frac{p^2}{1-z}+\frac{q^2}{z}+2bw-b^2\\
			\text{s.t. }& w-b = p-q,\;p\ge0,\;q\ge0.
		\end{align*}
		It can be seen easily that at the optimal solution, $p=(w-b)_+$ and $q=(w-b)_-$. The conclusion follows by substituting out the optimal $p,q$.
	\end{itemize}
	This finishes the proof.
\end{proof} 
We can use, using an appropriate linear transformation of variables, recover the proof of Theorem~\ref{theo:convexHull} from Lemma~\ref{lem:cvxHull}.
\begin{proof}[Proof of Theorem~\ref{theo:convexHull}]
	Without loss of generality, we assume $y=1$; otherwise one can change variables $x\gets yx$. Denote $\bm U$ as an orthonormal matrix whose first column is $\bm Q^{-1/2}\bm x/ \norm{2}{\bm Q^{-1/2}\bm x}$. Note that $\bm U^{-1}=\bm U^\top$. Let $\bm u= \bm U^\top \bm Q^{1/2} \bm w$. By the construction, one has $\bm w^\top \bm Q\bm w=\norm{2}{\bm U\bm u}^2=\norm{2}{\bm u}^2$, Moreover, $\bm x^\top \bm w=\bm x^\top \bm Q ^{-1/2}\bm U\bm u=\big(\bm U^\top \bm Q^{-1/2}\bm x\big)^\top \bm u=\norm{2}{\bm Q^{-1/2}\bm x}u_1$. Let $b = 1/\norm{2}{\bm Q^{-1/2}\bm x}$. The change of variables leads to the following bijective affine transformation of $W_Q$, which is denoted by $\widetilde W$
	\begin{align*}
		\widetilde W = \Big\{(\bm u, z, t):\;&t\ge \norm{2}{\bm u}^2, \; z\in\{0,1\},
		(b-u_1)(1-z)\le0,\; (b-u_1)z\ge 0\Big\}. 
	\end{align*}
	Since $\norm{2}{\bm u}^2=\sum_{i=1}^pu_i^2$ is separable in $u_i$, and $u_i$ is only involved in the constraint $t\ge\norm{2}{\bm u}^2$ for all $i\ge 2$, one can deduce that 
	\begin{align*}
		&\conv(\widetilde W)=\left\{(\bm u,z,t):\left(u_1,z,t-\sum_{i=2}^2u_i^2\right)\in\conv(\widehat W) \right\}  \\
		&= \bigg\{ (\bm u,z,t):\; 0\le z\le 1,
		\;t\ge \sum_{i=2}^p u_i^2+\frac{(b-u_1)_-^2}{1-z}+\frac{(b-u_1)_+^2}{z}+2bu_1-b^2 \bigg\}.
	\end{align*}
	The formula for the convex hull of $W$ can be recovered by substituting out $\norm{2}{\bm u}^2=\bm w^\top \bm Q\bm w$, $b-u_1=1/\norm{2}{\bm Q^{-1/2}\bm x}-\bm x^\top\bm w/\norm{2}{\bm Q^{-1/2}\bm x}=\sqrt{\gamma}\left(1-\bm x^\top \bm w\right)$:
	\begin{align*}
		t\ge&\sum_{i=2}^p u_i^2+\frac{(b-u_1)_-^2}{1-z}+\frac{(b-u_1)_+^2}{z}+2bu_1-b^2\\
		=& \norm{2}{\bm u}^2+\frac{(b-u_1)_-^2}{1-z}+\frac{(b-u_1)_+^2}{z}-(b-u_1)^2\\
		=& \bm{w^\top Qw}-\gamma\Big(1-(\bm{x^\top w})\Big)^2
		+\gamma\frac{\Big(1-\bm{x^\top w}\Big)_+^2}{z}+\gamma\frac{\Big(1-\bm{x^\top w}\Big)_-^2}{1-z},
	\end{align*}
	thus concluding the proof.
\end{proof}

\subsection{Interpretation as regularization}

From Theorem~\ref{theo:convexHull}, we found a representation (with an additional variable $z_i$) of the best loss function based on a single margin $y_i\bm{x_i^\top w}$. To retrieve a representation in the original space of variables, it suffices to project out the additional variable:
\begin{align}\hspace{-0.7em} \mathcal{L}^*(y_i\bm{x_i^\top w};\gamma_i):=\min_{0\leq z_i\leq 1}\,\lambda z_i&-\gamma_i\Big(1-y_i\bm{x_i^\top w}\Big)^2\\
	&+\gamma_i\frac{\Big(1-y_i(\bm{x_i^\top w})\Big)_+^2}{z_i}+\gamma_i\frac{\Big(1-y_i\bm{x_i^\top w}\Big)_-^2}{1-z_i}\label{eq:maxminPenaltySep}\end{align}
\noindent for some properly chosen $\gamma_i> 0$. Observe that we remove the subscript from $\mathcal{L}$ since the dependence on the datapoint is captured by $\gamma_i$. The resulting function is given in Proposition~\ref{prop:loss} below. 

\ignore{
	First, to better understand the proposed relaxation, we show that \eqref{eq:maxmin2} can be interpreted as using a specific non-convex loss function to penalize points based on their distance to the decision boundary. Using functions $h_i$ defined in \eqref{eq:defH}, we can rewrite \eqref{eq:maxmin2} as
	\begin{equation}\label{eq:maxminPenalty}
		\max_{\gamma\in \R_+^n:\eqref{eq:psd}}\;\min_{\substack{\bm{w}\in \R^p\\ \bm{0}\leq \bm{z}\leq \bm{1}}} \|\bm{w}\|_2^2+\sum_{i=1}^n \gamma_ih_i(\bm{w},z_i)+\lambda \sum_{i=1}^n z_i.
	\end{equation}
	Observe that, given some values for $\bm{\gamma}$ and $\bm{w}$, problem \eqref{eq:maxminPenalty} is separable on misclassifications variables $z_{i}$ for each datapoint $i\in [n]$. We focus on finding the solutions for $z_i$ by optimizing the following problem 
	\begin{equation}\label{eq:maxminPenaltySep}
		L(\bm{w}):=\min_{0\leq z_i\leq 1} \gamma_ih_i(\bm{w},z_i)+\lambda z_i.
	\end{equation}
}
\begin{proposition}\label{prop:loss}
	Assuming $\gamma_i,\lambda>0$, the optimal value of problem \eqref{eq:maxminPenaltySep} is the non-convex loss function 
	\begin{align*}
		\mathcal{L}^*(u;\gamma_i) = \begin{cases} 0 &\text{if}\ \ 1-u\leq 0 \\ 2\sqrt{\lambda\gamma_i}(1-u)-\gamma_i(1-u)^2 &\text{if}\ \ 0<1-u\leq \sqrt{\lambda/\gamma_i}\\ \lambda &\text{if}\ \ 1-u>\sqrt{\lambda/\gamma_i}  \end{cases}
	\end{align*}\normalsize
	where $u=y_i\bm{x_i^\top w}$.
\end{proposition}
\begin{proof}
	To determine the optimal value of $z_i$ in \eqref{eq:maxminPenaltySep}, we consider two cases depending on the value of $y_i(\bm{x_i^\top w})$.
	
	\noindent $\bullet$ \emph{Case 1: $1-y_i(\bm{x_i^\top w})\leq 0$.} In this case, the point is classified on the correct side of the margin. We find that the optimal value is $z_i^*=0$ as the objective is minimized to zero.  
	
	\noindent$\bullet$ \emph{Case 2: $1-y_i(\bm{x_i^\top w})\geq 0$.} In this case, we find by setting the derivative with respect to $z_i$ to zero  \eqref{eq:maxminPenaltySep} that the optimal solution is 
	$$z_i^*=\min\left\{\sqrt{\frac{\gamma_i}{\lambda}}\Big(1-y_i(\bm{x_i^\top w})\Big),1\right\},$$
	with objective value $2\sqrt{\lambda\gamma_i}(1-y_i(\bm{x_i^\top w}))-(1-y_i(\bm{x_i^\top w}))^2$ if $0<1-y_i(\bm{x_i^\top w})\leq \sqrt{\lambda/\gamma_i}$, and $\lambda$ if $1-y_i(\bm{x_i^\top w})>\sqrt{\lambda/\gamma_i}$.
\end{proof}

Figure~\ref{fig:lossFunctions} depicts the loss function $\mathcal{L^*}$ along with alternatives from the literature. The parameter $\gamma_i$ controls the tradeoff between non-convexity and approximation to the 0-1 loss. Note that for $\gamma_i=1$, the derived loss is a better approximation of the misclassification loss than the normalized sigmoid, $\psi$-learning, and ramp losses. In problems with uncertainty and outliers, the concave downward shape of the loss function redistributes the influence from outliers to points near the boundary. This reduces the sensitivity to noise when compared to hinge and other non-convex losses.  

Note that using loss function $\mathcal{L}^*$ in practice is not trivial. First, it requires a careful selection of parameters $\bm{\gamma}$ to guarantee an ideal tradeoff between convexity and strength. While Theorem~\ref{theo:convexHull} gives the best value in the context of a single observation $(\bm{x_i},y_i)$, it is not immediately clear how to generalize the result. Indeed, if all $\gamma_i$ are simultaneously set according to Theorem~\ref{theo:convexHull}, i.e. $\gamma_i=1/\|\bm{x_i}\|_2^2$, then the resulting problem is non-convex, posing difficulties for optimization. Second, even if $\bm{\gamma}$ is given, optimization with the loss function as described in Proposition~\ref{prop:loss} can be difficult, as it is non-smooth and defined by pieces (thus not directly compatible with off-the-shelf solvers). In the next section we discuss how to resolve these issues. 

\section{Implementation via conic optimization}\label{sec:primal}
Consider now problem \eqref{eq:separable} where each loss function corresponds to \eqref{eq:maxminPenaltySep} for suitable values of $\gamma_i$, that is, 
\begin{align}\nu^*=\min_{\substack{\bm{w}\in \R^p\\\bm{z}\in [0,1]^n}}\;&\|\bm{w}\|_2^2+\lambda\sum_{i=1}^nz_i-\sum_{i=1}^n\gamma_i\left(1-y_i\bm{x_i^\top w}\right)^2\\
	&+\sum_{i=1}^n\gamma_i\left(\frac{\left(1-y_i\bm{x_i^\top w}\right)_+^2}{z_i}+\frac{\left(1-y_i\bm{x_i^\top w}-1\right)_-^2}{1-z_i}\right).\label{eq:formulationLoss}
\end{align}\normalsize
Note that if constraints $\bm{z}\in \{0,1\}^n$ were additionally imposed, \eqref{eq:formulationLoss} is a valid formulation for problem \eqref{eq:BigMPenalty} , provided that $\bm{\gamma}>\bm{0}$; moreover, the relaxation is stronger than the big-M/hinge relaxation.

Since convex quadratic terms divided by nonnegative linear terms are convex and second-order cone representable \cite{Alizadeh2003,Lobo1998}, it follows that the terms in the last summation of \eqref{eq:formulationLoss} are convex. Thus, to ensure convexity of the objective in \eqref{eq:formulationLoss}, it suffices to ensure the convexity of the
function $\|\bm{w}\|_2^2-\sum_{i=1}^n\gamma_i\left(1-y_i\bm{x_i^\top w}\right)^2$, which is equivalent to
\begin{equation}\label{eq:psd}
	\bm{I}-\sum_{i=1}^n\gamma_i\bm{x_ix_i^\top}\succeq 0.\end{equation}
Therefore, the relaxation of SVM we propose can be formulated as the max/min optimization problem
\begin{align}
	\max_{\bm\gamma\in \R_+^n:\eqref{eq:psd}}\;\min_{\substack{\bm{w}\in \R^p\\\bm{z}\in [0,1]^n}}\;&\|\bm{w}\|_2^2+\lambda\sum_{i=1}^nz_i-\sum_{i=1}^n\gamma_i\left(1-y_i\bm{x_i^\top w}\right)^2\\
	&+\sum_{i=1}^n\gamma_i\left(\frac{\left(1-y_i\bm{x_i^\top w}\right)_+^2}{z_i}+\frac{\left(1-y_i\bm{x_i^\top w}\right)_-^2}{1-z_i}\right),\label{eq:maxmin}
\end{align}
where the inner minimization solves the convex relaxation of the MIO problem, and the outer maximization seeks to find vector $\bm{\gamma}$ resulting in the strongest convex relaxation. 

We start by denoting $h_i(\bm{w}, z_i)\geq 0$ as the strengthening associated with point $i$
\begin{align}
	\hspace{-0.8em}h_i(\bm{w}, z_i)=\left(\frac{(1-y_i(\bm{x_i^\top w}))_+^2}{z_i}+\frac{(1-y_i(\bm{x_i^\top w}))^2_-}{1-z_i}\right)-(1-y_i(\bm{x_i^\top w)})^2.\label{eq:defH}
\end{align}
Observe that function $h_i$ corresponds to $\mathcal{L}_i^*$ with $\gamma_i=1$, but we make explicit the dependence on the additional variable $z_i$.
In addition of variables $\bm{0}\leq \bm{z}\leq \bm{1}$ and $\bm{w}\in \R^p$, we will introduce a matrix variable $\bm{W}\in \R^{p\times p}$ which can be interpreted as representing the outer product $\bm{W}\approx \bm{ww^\top}$. 
\begin{theorem}\label{theo:conicPrimal}
	Problem \eqref{eq:maxmin} is equivalent to the SDP 
	\begin{subequations}\label{eq:primal}
		\begin{align}
			\min_{\bm{w},\,\bm{z},\, \bm{W}}\;&\trace(\bm{W})+\lambda\sum_{i=1}^nz_i\\
			\text{s.t. }&\langle\bm{x_i}\bm{x_i}^\top, \bm{W}\rangle-2y_i(\bm{x_i^\top w})+1\geq\frac{(1-y_i(\bm{x_i^\top w}))_+^2}{z_i}\\&\hphantom{\langle\bm{x_i}\bm{x_i}^\top, \bm{W}\rangle-2y_i(\bm{x_i^\top w)}+1\geq\;}+\frac{(1-y_i(\bm{x_i^\top w}))_-^2}{1-z_i}\quad\forall i\in [n]\\
			&\begin{pmatrix}1&\bm{w^\top}\\\bm{w}&\bm{W}\end{pmatrix}\succeq 0, \;\bm{w}\in \R^p,\;\bm{W}\in \R^{p\times p}\\
			&\bm z\in[0,1]^n.
		\end{align}
	\end{subequations}
\end{theorem}
\begin{proof}
	We first interchange $\max$ and $\min$ in \eqref{eq:maxmin}. Note that the objective is convex in $\bm w$ and $\bm z$, and linear in $\bm\gamma$. Moreover, the feasible region of $\bm\gamma$ is compact and convex as it is bounded by the positive semidefinite \eqref{eq:psd} and linear constraints. Thus, using Sion’s Minimax Theorem \cite{Sion58}, we switch the order of the operators, and \eqref{eq:maxmin} reduces to
	\begin{subequations}
		\begin{align}
			\min_{\bm{w},\,\bm{0}\leq\bm{z}\leq \bm{1}}\;\max_{\bm{\gamma},\,\bm{A}}\;&\|\bm{w}\|_2^2+\lambda\sum_{i=1}^n z_i+\sum_{i=1}^n \gamma_ih_i(\bm{w},z_i)\notag\\
			\text{s.t.}\;&\bm{I}-\sum_{i=1}^n \gamma_i\bm{x_ix_i^\top}=\bm{A}\tag{$\bm{V}$}\\
			&\bm{\gamma}\geq \bm{0}\tag{$\bm{s}$}\\
			&\bm{A}\succeq \bm{0}\tag{$\bm{S}$}.\notag
		\end{align}
	\end{subequations}
	Next, we dualize the inner maximization problem. The dual variables associated with each constraint are introduced in the above formulation. We obtain the conic dual
	\begin{align*}
		\min_{\substack{\bm{w}\in \R^p,\,\bm{0}\leq\bm{z}\leq \bm{1}\\\bm{s}\geq 0,\,\bm{S}\succeq 0}}\|\bm{w}\|_2^2+\lambda\sum_{i=1}^nz_i +\max_{\bm{\gamma},\,\bm{A}}\;&\left\{\sum_{i=1}^n \gamma_ih_i(\bm{w},z_i)
		+\bm{s^\top \gamma}
		+\langle \bm{S},\bm{A}\rangle \vphantom{\left\langle\bm{V},\bm{A}+\sum_{i=1}^n\gamma_i\bm{x_ix_i^\top}-\bm{I}\right\rangle }\right.\\
		&\quad\;\left.+\left\langle\bm{V},\bm{A}+\sum_{i=1}^n\gamma_i\bm{x_ix_i^\top}-\bm{I}\right\rangle\right\}.
	\end{align*}
	Observe that the inner maximization problem is unbounded unless $-s_i=h_i(\bm{w},z_i)+\langle \bm{V},\bm{x_ix_i^\top}\rangle$ for all points, and $-\bm V=\bm A$. Thus, the problem reduces to
	\begin{align*}
		\min_{\bm w,\,\bm z,\,\bm V}\;&\|\bm{w}\|_2^2+\lambda\sum_{i=1}^nz_i-\langle\bm{V},\bm{I}\rangle\\
		\text{s.t.}\;& h_i(\bm{w},z_i)+\bm{x_i^\top Vx_i}\leq 0\quad \forall i\in [n]\\
		&-\bm{V}\succeq 0,\,\bm{V}\in \R^{p\times p}\\
		&\bm{w}\in \R^p,\,\bm{0}\leq\bm{z}\leq \bm{1}.
	\end{align*}
	Finally, we perform the change of variables $\bm{W} = \bm{ww^\top-V}$, we obtain the formulation
	\begin{align*}
		\min_{\bm w,\,\bm z,\,\bm W}\;&\sum_{j=1}^n W_{jj}+\lambda\sum_{i=1}^nz_i\\
		\text{s.t.}\;& h_i(\bm{w},z_i)+(\bm{x_i^\top w})^2-\langle\bm {x_ix_i^\top}, \bm{W}\rangle\leq 0 \quad \forall i\in[n]\\
		&\bm{W}\succeq \bm{ww^\top},\,\bm{w}\in \R^p,\,\bm{W}\in \R^{p\times p}\\
		&\bm{0}\leq\bm{z}\leq \bm{1}.
	\end{align*}
	We recover \eqref{eq:primal} by replacing $h_i$ with its definition and using the Schur complement for constraints $\bm{W-ww^\top}\succeq 0$. 
\end{proof}

Note that problem \eqref{eq:primal} is an SDP with $n+p+p^2$ variables, $n$ constraints and positive definite constraints involving cones of order $p+1$. We point out that while SDPs are infamous for being difficult to solve, the main factor affecting runtimes with leading conic solvers --in our case, Mosek-- is the order of the positive semidefinite cones.  In particular, in our computations presented in the next section, we observe that runtimes scale almost linearly with the number of datapoints $n$, but polynomially with the number of features $p$. Thus, the proposed approach can be directly used is datasets with $n$ in the thousands, provided that the number of features is relatively small -- these types of datasets are common in high-stakes domains such as policy-making and healthcare, e.g., see \cite{rudin2022black} and the references therein.

\begin{remark}
	In practice, Kernel transformations are often used to produce classifiers that are nonlinear. The proposed approach extends to Kernel formulations, and we discuss this extension in Appendix~\ref{sec:kernel}.
\end{remark}

\section{Computations}\label{sec:computations}

We now discuss computations with the proposed methods. 
\subsection{Synthetic instances}
We describe in \S\ref{sec:instance} the instance generation process, in \S\ref{sec:methods} the methods used and present in \S\ref{sec:resultsSynt} results.

\subsubsection{Instance generation}\label{sec:instance}
The instances considered are inspired by the instances in \cite{brooks2011support}: the data from each class is generated from a different Gaussian distribution, and the presence of different types of outliers may obfuscate the data. We now describe in detail the data generation process.

Specifically, let $n,p\in \Z_+$ be dimension parameters, and $\sigma\in \R_+$ represent the standard deviation of the Gaussian distributions. We consider three outlier classes (none, clustered, spread). We first generate a direction $\bm{d}\in \R^p$ where each entry is generated independently from a uniform distribution in $[-1,1]$. We then set two centroids $\bm{\chi_1}=0.5\bm{d}/\|\bm{d}\|_2$ and $\bm{\chi_{-1}}=-0.5\bm{d}/\|\bm{d}\|_2$, and note that they are always one unit apart. Then we generate $n$ points $\bm{x_i}\in \R^{p+1}$ where $(x_i)_1=1$ and the remaining coordinates depend on the outlier class, and are generated as follows:

\noindent$\bullet$ \textbf{none}  With $0.5$ probability the remainder coordinates are generated from $\mathcal{N}(\chi_1,\sigma^2\bm{I})$ and set the label $y_i=1$, and with $0.5$ probability the remainder coordinates are generated from $\mathcal{N}(\chi_{-1},\sigma^2\bm{I})$ and set the label $y_i=-1$.

\noindent$\bullet$ \textbf{clustered} With $0.45$ probability the remainder coordinates are generated from $\mathcal{N}(\chi_1,\sigma^2\bm{I})$ and set the label $y_i=1$, with $0.45$ probability the remainder coordinates are generated from $\mathcal{N}(\chi_{-1},\sigma^2\bm{I})$ and set the label $y_i=-1$, and with $0.1$ probability the remainder coordinates are generated from $\mathcal{N}(10\chi_{-1},0.001\sigma^2\bm{I})$ and set the label $y_i=1$.

\noindent$\bullet$ \textbf{spread} An expected 45\% of the points are generated from $\mathcal{N}(\chi_1,\sigma^2\bm{I})$ and set the label $y_i=1$, 45\% are generated from $\mathcal{N}(\chi_{-1},\sigma^2\bm{I})$ and set the label $y_i=-1$, 5\% of the points are generated from $\mathcal{N}(\chi_1,100\sigma^2\bm{I})$ and set the label $y_i=1$ and 5\% of the points are generated from $\mathcal{N}(\chi_{-1},100\sigma^2\bm{I})$ and set the label $y_i=-1$. 
In all cases, the \emph{ideal} Bayes classifier is the line perpendicular to $\bm{\chi_1}-\bm{\chi_{-1}}$ going through the origin, $\bm{\hat w^\top}=(0\; \bm{d^\top})$. We point out that we performed additional experiments where the data is perfectly separable but some labels are flipped: the conclusions are similar to the ones presented here (and also our results with real data), hence we defer the presentation of these computations to Appendix~\ref{sec:compAdditional}.

\subsubsection{Methods, metrics and implementation}\label{sec:methods}
We compare the following three methods: SVMs with the hinge loss, SVMs with the conic loss, and the Bayes classifier described in the previous subsubsection (corresponding to the best possible performance). In each experiment, we generate a training and validation set (both contaminated by outliers for classes ``clustered" and ``spread"), each with $n$ observations, and a testing set without outliers and $100,000$ observations. For the SVM approaches, we solve the training problem on the training set for 100 different values of the hyperparameter, choose the solution that results in fewer misclassified points in validation, and report the (out-of-sample) misclassification rate on the testing set, as well as the total time required to perform cross-validation (i.e., solving 100 training problems). For each combination of instance class and dimensions, we repeat this process 20 times and report averages and standard deviations across all replications. For SVMs with the hinge loss, we set $\lambda=\beta/(1-\beta)$ where the 100 values of $\beta$ are selected uniformly in the interval $(0,1)$; for SVMs with the conic loss, we use the method described in Appendix~\ref{sec:crossvalidation}.

To train SVMs with the conic loss, we use Mosek 10.0 solver (with default parameters) on a laptop with a 12th Gen Intel Core i7-1280P (20 CPUs) processor and 32 GM RAM. For the hinge loss, we solve the training problems as convex quadratic programs using Gurobi 10.0.1 solver (default parameters) on the same laptop.

\subsubsection{Results}\label{sec:resultsSynt}
\begin{figure}[!h]
	\centering
	\subfloat[${n=100}$, \newline 
	${\text{hinge}=17.0\%\pm9.2\%}$,
	$\text{conic}=1.9\%\pm2.0\%$]{\includegraphics[width=0.45\columnwidth,trim={13cm 6cm 14cm 6cm},clip]{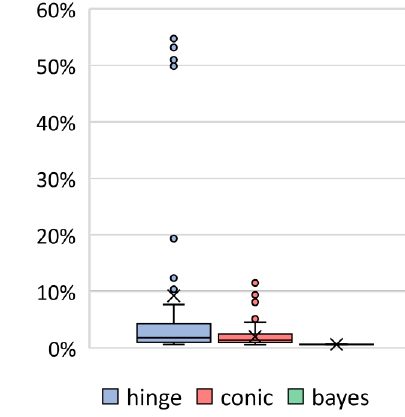}}\hfill
	\subfloat[${n=200}$,\newline ${\text{hinge}=12.0\%\pm4.6\%}$, $\text{conic}=1.1\%\pm1.3\%$]{\includegraphics[width=0.45\columnwidth,trim={13cm 6cm 14cm 6cm},clip]{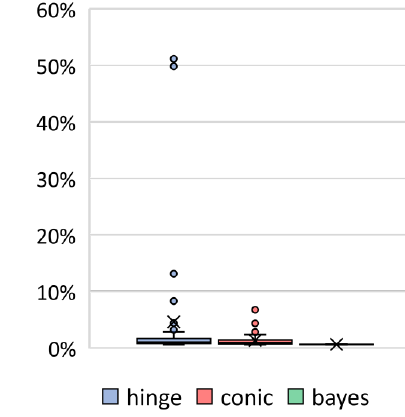}}\hfill
	\newline
	\subfloat[${n=500}$, \newline
	${\text{hinge}=0.9\%\pm0.3\%}$,	$\text{conic}=0.8\%\pm0.2\%$]{\includegraphics[width=0.45\columnwidth,trim={13cm 6cm 14cm 6cm},clip]{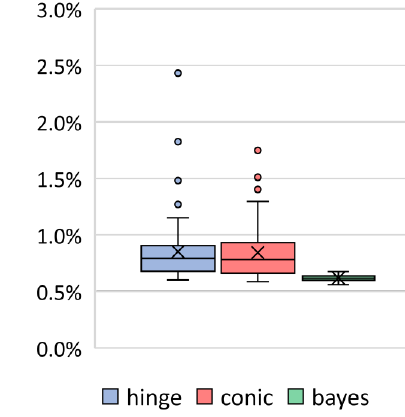}}\hfill
	\subfloat[${n=1,000}$, \newline
	${\text{hinge}=0.7\%\pm0.1\%}$,
	$\text{conic}=0.8\%\pm0.2\%$]{\includegraphics[width=0.45\columnwidth,trim={13cm 6cm 14cm 6cm},clip]{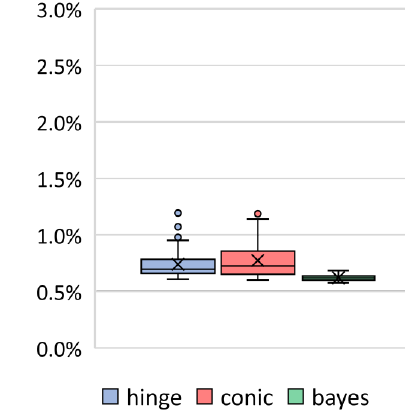}}
	\caption{\small Distribution of out-of-sample misclassification for data with clustered outliers and $\sigma=0.2$, as a function of the number of datapoints $n$. In instances with small $n$ (top row), the hinge estimator has a probability of breaking down, resulting in out-of-sample misclassifications above 50\%; \emph{the conic loss reduces the average misclassification rate by an order-of-magnitude}. Moreover, the conic estimator performs consistently good in all settings.}
	\label{fig:oos}
\end{figure}

Table~\ref{tab:misclassification2Synt} shows the results in synthetic instances. We observe that in regimes with no outliers, SVMs with the hinge and conic losses result in almost identical performance of the ensuing estimators. However, in regimes with outliers, using the conic loss typically results in superior performance, having better average out-of-sample misclassification \emph{and} smaller variance. This phenomenon is particularly visible in instances with ``clustered" outliers and $\sigma=0.2$ (shown in bold in Table~\ref{tab:misclassification2Synt}), and Figure~\ref{fig:oos} depicts detailed results in this setting. We observe that the large variance of the hinge estimator is due to a subset of instances where the estimator breaks down, resulting in solutions with over 50\% misclassification rate. In contrast, the conic loss does not exhibit such a pathological behavior and performs well consistently.

\begin{table}[!h]
	\renewcommand{\arraystretch}{1.2}
	\caption{\small Out-of-sample misclassification rate with synthetic instances, as a function of outlier generation. Results in \textbf{bold} (clustered, $\sigma=0.2$) are presented in more detail in Figure~\ref{fig:oos}.} 
	\label{tab:misclassification2Synt}
	\setlength{\tabcolsep}{2pt}
	\begin{tabular}{ c c | c |c |c }
		\hline
		$\bm{\sigma}$&\textbf{method}&\textbf{none}&\textbf{clustered}&\textbf{spread}\\
		\hline
		\multirow{3}{*}{0.2}&hinge&$0.8\%\pm 0.2\%$&$\bm{3.9\%\pm 10.9\%}$&$0.9\%\pm 0.5\%$\\
		&conic&$1.0\%\pm 0.5\%$&$\bm{1.2\%\pm 1.2\%}$&$1.0\%\pm 0.6\%$\\
		&bayes&$0.6\%\pm 0.0\%$&$\bm{0.6\%\pm 0.0\%}$&$0.6\%\pm 0.0\%$\\
		\hline
		\multirow{3}{*}{0.5}&hinge&$16.8\%\pm 1.3\%$&$21.7\%\pm 8.7\%$&$18.2\%\pm 2.3\%$\\
		&conic&$17.0\%\pm 1.3\%$&$18.3\%\pm 3.5\%$&$17.5\%\pm 1.8\%$\\
		&bayes&$15.9\%\pm 0.1\%$&$15.9\%\pm 0.1\%$&$15.9\%\pm 0.1\%$\\
		\hline
		\multirow{3}{*}{1.0}&hinge&$32.5\%\pm 1.9\%$&$36.5\%\pm 5.6\%$&$34.4\%\pm 3.0\%$\\
		&conic&$32.6\%\pm 1.9\%$&$34.1\%\pm 3.4\%$&$32.9\%\pm 2.1\%$\\
		&bayes&$30.9\%\pm 0.1\%$&$30.9\%\pm 0.1\%$&$30.9\%\pm 0.1\%$\\
		\hline		
	\end{tabular}
\end{table}

Finally, Table~\ref{tab:solTimesSynt} presents the computational times required to select a model using cross-validation as described before. We found that the computational times depend on $n$ and $p$ but not on the instance class, thus we present aggregated results across all instance classes. We observe that while the conic loss is certainly more expensive to solve than the hinge loss, instances with up to $n=1,000$ are solved in less than a second (and thus the complete cross-validation process requires less than 100 seconds).

\begin{table}[h]
	\renewcommand{\arraystretch}{1.2}\vspace{-0.5em}
	\caption{\small Average solution times and standard deviations (in seconds) across all instance classes required to solve the 100 training problems for cross-validation. } 
	\label{tab:solTimesSynt}
	\setlength{\tabcolsep}{2pt}
		\begin{tabular}{c c | c |c |c |c}
			\hline
			$\bm{n}$&\textbf{method}&$\bm{p=3}$&$\bm{p=5}$&$\bm{p=10}$&$\bm{p=30}$\\
			\hline
			\multirow{2}{*}{$\bm{100}$}&hinge&$0.2\pm 0.0$ &$0.2\pm 0.0$&$0.2\pm 0.0$&$0.3\pm 0.0$\\
			&conic&$2.8\pm 0.3$&$2.9\pm 0.4$&$5.2\pm 1.4$&$41.2\pm 4.7$\\
			\hline
			\multirow{2}{*}{$\bm{200}$}&hinge&$0.5\pm 0.1$ &$0.5\pm 0.1$&$1.9\pm 0.2$&$2.2\pm 2.0$\\
			&conic&$3.5\pm 0.5$&$3.7\pm 0.5$&$8.7\pm 2.6$&$94.1\pm 42.7$\\
			\hline
			\multirow{2}{*}{$\bm{500}$}&hinge&$6.2\pm 1.4$ &$6.2\pm 1.5$&$5.8\pm 0.8$&$11.3\pm 2.0$\\
			&conic&$14.7\pm 4.7$&$14.5\pm 4.7$&$20.6\pm 4.9$&$152.8\pm 25.8$\\
			\hline
			\multirow{2}{*}{$\bm{1,000}$}&hinge&$0.6\pm 0.1$ &$0.7\pm 0.1$&$1.1\pm 0.1$&$2.0\pm 0.4$\\
			&conic&$18.5\pm 2.3$ &$20.5\pm 2.6$&$25.2\pm 9.3$&$167.2\pm 62.5$\\
			\hline
		\end{tabular}\vspace{-0.5em}
\end{table}

\subsection{Real instances}
We test the proposed methods in five datasets from the UCI machine learning repository \citep{Dua:2019}, whose names and dimensions (along with computational results) are reported in Table~\ref{tab:resultsReal}. We split each dataset randomly with 35\% of the observations in a training set, 35\% in a validation set and 30\% in the testing set. Additionally, given a noise parameter $\tau\in [0,0.5)$, we introduce outliers by flipping the label of each observation of the training and validation set with probability $\tau$ (the test set is not corrupted by outliers). We then perform cross-validation for the hinge and conic loss as described in \S\ref{sec:methods}. For each dataset and value of the noise parameter $\tau$, we perform 20 random splits of train/validation/test and report average results and standard deviations over these replications.

While the results are highly dependent on the specific dataset (e.g., hinge is better across all noise parameters for the sonar dataset; conic is better across all parameters for the breast cancer dataset), we observe that on average SVMs with the hinge loss perform better in instances without outlier $\tau=0$, while SVMs with the conic loss are better in settings with outliers $(\tau\geq 0.2)$. The results are thus consistent with those observed for synthetic data. Moreover, we also comment that, regardless of the value of the noise parameter $\tau$, using the conic loss results in smaller standard deviations. As discussed in Figure~\ref{fig:oos}, large standard deviations indicate poor performance in some instances, thus the smaller standard deviations suggest that the conic loss results in estimators that are more robust to the specific selection of training/validation/testing sets.

\begin{landscape}
	\begin{table}[!h t b]
		\renewcommand{\arraystretch}{1.5}
		\centering
		\caption{\small Results with real datasets, where $\tau$ is the probability of a label swap and each row shows averages and standard deviations over 20 splits of training/validation/testing. Column ``time" represents the total time required solve 100 SVM problems (required for cross-validation). We depict in bold the best average out-of-sample accuracy for each dataset-noise combination.}
		\label{tab:resultsReal}
		\setlength{\tabcolsep}{2pt}
		\resizebox{0.9\linewidth}{!}{%
			\begin{tabular}{c c c c | c | c c  c c}
				\hline
				\multirow{2}{*}{\textbf{name}}&\multirow{2}{*}{\textbf{n}}&\multirow{2}{*}{\textbf{p}}&\multirow{2}{*}{\textbf{method}}&\multirow{2}{*}{\textbf{time (s)}}&\multicolumn{4}{c}{\underline{\textbf{out-of-sample misclassification rate}}}\\
				&&&&&$\bm{\tau=0.0}$&$\bm{\tau=0.1}$&$\bm{\tau=0.2}$&$\bm{\tau=0.3}$\\
				
				\hline
				\multirow{2}{*}{Breast cancer}&\multirow{2}{*}{196}&\multirow{2}{*}{36}&hinge&$0.1\pm 0.0$&$16.9\%\pm 4.4\%$&$18.3\%\pm 4.6\%$&$17.9\%\pm 5.0\%$&$19.4\%\pm 7.7\%$\\
				&&&conic&$45.4\pm 4.1$&$\bm{16.8\%\pm 4.2\%}$&$\bm{17.6\%\pm 4.5\%}$&$\bm{17.5\%\pm 4.4\%}$&$\bm{18.8\%\pm 5.5\%}$\\
				\hline
				\multirow{2}{*}{German credit}&\multirow{2}{*}{1,000}&\multirow{2}{*}{24}&hinge&$3.1\pm 0.3$&$\bm{25.0\%\pm 1.9\%}$&$\bm{25.8\%\pm 2.3\%}$&$\bm{27.5\%\pm 2.0\%}$&$\bm{29.4\%\pm 2.4\%}$\\
				&&&conic&$72.0\pm 6.2$&$29.6\%\pm 2.1\%$&$29.4\%\pm 2.3\%$&$29.7\%\pm 2.2\%$&$29.5\%\pm 1.9\%$\\
				\hline
				\multirow{2}{*}{Immunotherapy}&\multirow{2}{*}{90}&\multirow{2}{*}{6}&hinge&$0.1\pm 0.0$&$\bm{20.5\%\pm 8.0\%}$&$24.1\%\pm 9.1\%$&$28.6\%\pm 10.4\%$&$34.6\%\pm 11.9\%$\\
				&&&conic&$2.5\pm 0.1$&$20.7\%\pm 6.3\%$&$\bm{20.7\%\pm 6.3\%}$&$\bm{21.1\%\pm 6.0\%}$&$\bm{27.0\%\pm 16.6\%}$\\
				\hline
				\multirow{2}{*}{Ionosphere}&\multirow{2}{*}{351}&\multirow{2}{*}{34}&hinge&$0.5\pm 0.0$&$16.0\%\pm 3.3\%$&$19.2\%\pm 4.4\%$&$20.9\%\pm 5.0\%$&$\bm{24.0\%\pm 6.9\%}$\\
				&&&conic&$82.0\pm 6.0$&$\bm{15.8\%\pm 3.1\%}$&$\bm{18.0\%\pm 3.6\%}$&$\bm{20.5\%\pm 4.8\%}$&$24.1\%\pm 6.4\%$\\
				\hline
				\multirow{2}{*}{Sonar}&\multirow{2}{*}{208}&\multirow{2}{*}{60}&hinge&$0.3\pm 0.0$&$\bm{25.4\%\pm 7.3\%}$&$\bm{30.0\%\pm 7.6\%}$&$\bm{32.7\%\pm 7.5\%}$&$\bm{42.3\%\pm 10.7\%}$\\
				&&&conic&$202.8\pm 15.1$&$30.1\%\pm 6.4\%$&$31.9\%\pm 6.1\%$&$37.3\%\pm 9.5\%$&$45.6\%\pm 6.8\%$\\
				\hline
				\multirow{2}{*}{Average}&&&hinge&-&$\bm{20.8\%\pm 5.0\%}$&$\bm{23.5\%\pm 5.6\%}$&$25.5\%\pm 6.0\%$&$29.9\%\pm 7.9\%$\\
				&&&conic&-&$22.6\%\pm 4.2\%$&$\bm{23.5\%\pm 4.6\%}$&$\bm{25.2\%\pm 5.4\%}$&$\bm{29.0\%\pm 7.4\%}$\\
				\hline
			\end{tabular}
		}
	\end{table}
\end{landscape}

\section{Conclusions} We developed an approach for training SVMs, by studying relaxations for mixed-integer optimization formulations of SVMs with the 0-1 loss. The resulting training problems are convex, amenable to solve using existing off-the-shelf solvers, and results in better performance (compared with the commonly used convex estimator involving the hinge loss) in settings with noise and outliers. 

\section*{Acknowledgments}

Andr\'es G\'omez is supported in part by grant FA9550-22-1-0369 from the Air Force Office of Scientific Research.

\bibliography{reference}

\begin{thebibliography}{}

\bibitem[Alizadeh and Goldfarb, 2003]{Alizadeh2003}
Alizadeh, F. and Goldfarb, D. (2003).
\newblock Second-order cone programming.
\newblock {\em Mathematical Programming}, 95:3--51.

\bibitem[Atamt\"urk and G\'omez, 2019]{atamturk2019rank}
Atamt\"urk, A. and G\'omez, A. (2019).
\newblock Rank-one convexification for sparse regression.
\newblock {\em arXiv preprint arXiv:1901.10334}.

\bibitem[Atamt\"urk and G{\'o}mez, 2020]{atamturk2020safe}
Atamt\"urk, A. and G{\'o}mez, A. (2020).
\newblock Safe screening rules for {L0}-regression from perspective
  relaxations.
\newblock In {\em International Conference on Machine Learning}, pages
  421--430. PMLR.

\bibitem[Bartlett et~al., 2006]{bartlett2006convexity}
Bartlett, P.~L., Jordan, M.~I., and McAuliffe, J.~D. (2006).
\newblock Convexity, classification, and risk bounds.
\newblock {\em Journal of the American Statistical Association},
  101(473):138--156.

\bibitem[Bertsimas et~al., 2023]{bertsimas2023sparse}
Bertsimas, D., Cory-Wright, R., and Johnson, N.~A. (2023).
\newblock Sparse plus low rank matrix decomposition: A discrete optimization
  approach.
\newblock {\em Journal of Machine Learning Research}, 24:1--51.

\bibitem[Bertsimas et~al., 2022]{bertsimas2022mixed}
Bertsimas, D., Cory-Wright, R., and Pauphilet, J. (2022).
\newblock Mixed-projection conic optimization: A new paradigm for modeling rank
  constraints.
\newblock {\em Operations Research}, 70(6):3321--3344.

\bibitem[Bertsimas and Van~Parys, 2020]{bertsimas2020sparse}
Bertsimas, D. and Van~Parys, B. (2020).
\newblock Sparse high-dimensional regression: Exact scalable algorithms and
  phase transitions.
\newblock {\em The Annals of Statistics}, 48(1):300--323.

\bibitem[Brooks, 2011]{brooks2011support}
Brooks, J.~P. (2011).
\newblock Support vector machines with the ramp loss and the hard margin loss.
\newblock {\em Operations Research}, 59(2):467--479.

\bibitem[Carrizosa and Morales, 2013]{carrizosa2013supervised}
Carrizosa, E. and Morales, D.~R. (2013).
\newblock Supervised classification and mathematical optimization.
\newblock {\em Computers \& Operations Research}, 40(1):150--165.

\bibitem[Cortes and Vapnik, 1995]{cortes1995support}
Cortes, C. and Vapnik, V. (1995).
\newblock Support-vector networks.
\newblock {\em Machine Learning}, 20:273--297.

\bibitem[d'Aspremont et~al., 2004]{d2004direct}
d'Aspremont, A., El~Ghaoui, L., Jordan, M., and Lanckriet, G. (2004).
\newblock A direct formulation for sparse {PCA} using semidefinite programming.
\newblock {\em Advances in Neural Information Processing Systems}, 17.

\bibitem[De~Rosa and Khajavirad, 2022]{de2022ratio}
De~Rosa, A. and Khajavirad, A. (2022).
\newblock The ratio-cut polytope and k-means clustering.
\newblock {\em SIAM Journal on Optimization}, 32(1):173--203.

\bibitem[Dey et~al., 2023]{dey2023solving}
Dey, S.~S., Molinaro, M., and Wang, G. (2023).
\newblock Solving sparse principal component analysis with global support.
\newblock {\em Mathematical Programming}, 199(1-2):421--459.

\bibitem[Dong et~al., 2015]{dong2015regularization}
Dong, H., Chen, K., and Linderoth, J. (2015).
\newblock Regularization vs. relaxation: A conic optimization perspective of
  statistical variable selection.
\newblock {\em arXiv preprint arXiv:1510.06083}.

\bibitem[Dua and Graff, 2017]{Dua:2019}
Dua, D. and Graff, C. (2017).
\newblock Uci machine learning repository.

\bibitem[Ghosh et~al., 2015]{ghosh2015making}
Ghosh, A., Manwani, N., and Sastry, P. (2015).
\newblock Making risk minimization tolerant to label noise.
\newblock {\em Neurocomputing}, 160:93--107.

\bibitem[G\'omez, 2021]{gomez2021outlier}
G\'omez, A. (2021).
\newblock Outlier detection in time series via mixed-integer conic quadratic
  optimization.
\newblock {\em SIAM Journal on Optimization}, 31(3):1897--1925.

\bibitem[Guan et~al., 2009]{guan2009mixed}
Guan, W., Gray, A., and Leyffer, S. (2009).
\newblock Mixed-integer support vector machine.
\newblock In {\em NIPS workshop on optimization for machine learning}.

\bibitem[G{\"u}nl{\"u}k and Linderoth, 2010]{Gunluk2010}
G{\"u}nl{\"u}k, O. and Linderoth, J. (2010).
\newblock Perspective reformulations of mixed integer nonlinear programs with
  indicator variables.
\newblock {\em Mathematical Programming}, 124:183--205.

\bibitem[Hazimeh et~al., 2022]{hazimeh2022sparse}
Hazimeh, H., Mazumder, R., and Saab, A. (2022).
\newblock Sparse regression at scale: Branch-and-bound rooted in first-order
  optimization.
\newblock {\em Mathematical Programming}, 196(1-2):347--388.

\bibitem[Kim et~al., 2022]{kim2022convexification}
Kim, J., Tawarmalani, M., and Richard, J.-P.~P. (2022).
\newblock Convexification of permutation-invariant sets and an application to
  sparse principal component analysis.
\newblock {\em Mathematics of Operations Research}, 47(4):2547--2584.

\bibitem[Li and Xie, 2020]{li2020exact}
Li, Y. and Xie, W. (2020).
\newblock Exact and approximation algorithms for sparse {PCA}.
\newblock {\em arXiv preprint arXiv:2008.12438}.

\bibitem[Lobo et~al., 1998]{Lobo1998}
Lobo, M.~S., Vandenberghe, L., Boyd, S., and Lebret, H. (1998).
\newblock Applications of second-order cone programming.
\newblock {\em Linear algebra and its applications}, 284:193--228.

\bibitem[Maldonado et~al., 2014]{maldonado2014feature}
Maldonado, S., P{\'e}rez, J., Weber, R., and Labb{\'e}, M. (2014).
\newblock Feature selection for support vector machines via mixed integer
  linear programming.
\newblock {\em Information sciences}, 279:163--175.

\bibitem[Manwani and Sastry, 2013]{manwani2013noise}
Manwani, N. and Sastry, P. (2013).
\newblock Noise tolerance under risk minimization.
\newblock {\em IEEE Transactions on Cybernetics}, 43(3):1146--1151.

\bibitem[Mason et~al., 1999]{mason1999boosting}
Mason, L., Baxter, J., Bartlett, P., and Frean, M. (1999).
\newblock Boosting algorithms as gradient descent.
\newblock {\em Advances in Neural Information Processing Systems}, 12.

\bibitem[Miller, 2002]{miller2002subset}
Miller, A. (2002).
\newblock {\em Subset selection in Regression}.
\newblock Chapman and Hall/CRC.

\bibitem[Peng and Xia, 2005]{peng2005new}
Peng, J. and Xia, Y. (2005).
\newblock A new theoretical framework for k-means-type clustering.
\newblock {\em Foundations and advances in data mining}, pages 79--96.

\bibitem[Pilanci et~al., 2015]{pilanci2015sparse}
Pilanci, M., Wainwright, M.~J., and El~Ghaoui, L. (2015).
\newblock Sparse learning via boolean relaxations.
\newblock {\em Mathematical Programming}, 151(1):63--87.

\bibitem[Rudin, 2022]{rudin2022black}
Rudin, C. (2022).
\newblock Why black box machine learning should be avoided for high-stakes
  decisions, in brief.
\newblock {\em Nature Reviews Methods Primers}, 2(1):81.

\bibitem[Shen et~al., 2003]{shen2003psi}
Shen, X., Tseng, G.~C., Zhang, X., and Wong, W.~H. (2003).
\newblock On $\psi$-learning.
\newblock {\em Journal of the American Statistical Association},
  98(463):724--734.

\bibitem[Sion, 1958]{Sion58}
Sion, M. (1958).
\newblock {On general minimax theorems.}
\newblock {\em Pacific Journal of Mathematics}, 8(1):171 -- 176.

\bibitem[Song et~al., 2002]{song2002robust}
Song, Q., Hu, W., and Xie, W. (2002).
\newblock Robust support vector machine with bullet hole image classification.
\newblock {\em IEEE transactions on systems, man, and cybernetics, part C
  (applications and reviews)}, 32(4):440--448.

\bibitem[Ustun et~al., 2013]{ustun2013supersparse}
Ustun, B., Traca, S., and Rudin, C. (2013).
\newblock Supersparse linear integer models for predictive scoring systems.
\newblock {\em arXiv preprint arXiv:1306.5860}.

\bibitem[Wu and Liu, 2007]{wu2007robust}
Wu, Y. and Liu, Y. (2007).
\newblock Robust truncated hinge loss support vector machines.
\newblock {\em Journal of the American Statistical Association},
  102(479):974--983.

\bibitem[Xie and Deng, 2020]{xie2020scalable}
Xie, W. and Deng, X. (2020).
\newblock Scalable algorithms for the sparse ridge regression.
\newblock {\em SIAM Journal on Optimization}, 30(4):3359--3386.

\bibitem[Xu et~al., 2017]{xu2017robust}
Xu, G., Cao, Z., Hu, B.-G., and Principe, J.~C. (2017).
\newblock Robust support vector machines based on the rescaled hinge loss
  function.
\newblock {\em Pattern Recognition}, 63:139--148.

\bibitem[Zhang, 2010]{zhang2010nearly}
Zhang, C.-H. (2010).
\newblock Nearly unbiased variable selection under minimax concave penalty.
\newblock {\em Annals of statistics}, 38(2):894--942.

\end{thebibliography}
\bibliographystyle{apalike}

\newpage
\appendix

\section{Kernel formulations}\label{sec:kernel}

Consider again the MIO formulation for SVM with the exact 0-1 loss:
\begin{subequations}\label{eq:BigMPenaltyRepeated}
	\begin{align}
		\min_{\bm{w},\,\bm{z}}\;&\|\bm{w}\|_2^2+\lambda\sum_{i=1}^n z_i\label{eq:BigMPenalty_obj}\\
		\text{s.t. }&\left(1-y_i\bm{x_i^\top w}\right)(1-z_i)\leq 0& \forall i\in [n]\\
		& \left(1-y_i\bm{x_i^\top w}\right)z_i\geq 0& \forall i\in [n]\\
		&\bm{w}\in \R^p,\,\bm{z}\in \{0,1\}^n.
	\end{align}
\end{subequations}

We now use the approach discussed in \cite{brooks2011support}, which involves the substitution $\bm{w}=\sum_{i=1}^n y_i\bm{x_i}\alpha_i$ for some variables  $\alpha_i$, resulting in the formulation
\begin{subequations}\label{eq:BigMKernel}
	\begin{align}
		\min_{\bm{\alpha},\,\bm{z}}\;&\sum_{i=1}^n\sum_{j=1}^ny_iy_j\bm{x_i^\top x_j}\alpha_i\alpha_j+\lambda\sum_{i=1}^n z_i\label{eq:BigMKernel_obj}\\
		\text{s.t.}\:&\left(1-\sum_{j=1}^ny_iy_j\bm{x_i^\top x_j}\alpha_j\right)(1-z_i)\leq 0&\forall i\in[n]\\ &\left(1-\sum_{j=1}^ny_iy_j\bm{x_i^\top x_j}\alpha_j\right)z_i\geq 0& \forall i\in [n]\\
		&\bm{\alpha}\in \R^n,\bm{z}\in \{0,1\}^n.
	\end{align}
\end{subequations}
The proof that formulations \eqref{eq:BigMPenaltyRepeated} and \eqref{eq:BigMKernel} are indeed equivalent can be found in \cite{brooks2011support}. The key idea is that inner products $\bm{x_i^\top x_j}$ can be replaced with $k(\bm{x_i},\bm{x_j})$ for a given kernel function $k:\R^p\times \R^p\to\R$, and if the kernel is positive-definite (as is the case with commonly used kernel functions) then the objective \eqref{eq:BigMKernel_obj} remains positive semidefinite.

The method proposed in the paper can be readily applied to formulation \eqref{eq:BigMKernel}. Indeed, the counterpart of formulation \eqref{eq:maxmin} is
\begin{equation}
	\resizebox{\textwidth}{!}{%
	$\displaystyle\begin{aligned}
		\max_{\bm\gamma\in \Gamma}\min_{\substack{\bm{\alpha}\in \R^n\\\bm{z}\in [0,1]^n}}&\;\sum_{i=1}^n\sum_{j=1}^ny_iy_jk(\bm{x_i},\bm{x_j})\alpha_i\alpha_j+\lambda\sum_{i=1}^nz_i-\sum_{i=1}^n\gamma_i\left(1-\sum_{j=1}^ny_iy_jk(\bm{x_i},\bm{x_j})\alpha_j\right)^2\\
		&\hspace{-2em}+\sum_{i=1}^n\gamma_i\left(\frac{\left(1-\sum_{j=1}^ny_iy_jk(\bm{x_i},\bm{x_j})\alpha_j\right)_+^2}{z_i}+\frac{\left(1-\sum_{j=1}^ny_iy_jk(\bm{x_i},\bm{x_j})\alpha_j\right)_-^2}{1-z_i}\right),\label{eq:maxminKernel}
	\end{aligned}$
}
\end{equation}
where $\Gamma\subseteq \R_+^n$ is the set that ensures that the quadratic term in the first line is positive semidefinite. Defining $\bm{K}\in \R^{n\times n}$ as the kernel matrix whose $(i,j)$-entry is $y_iy_jk(\bm{x_i},\bm{x_j})$, and letting $\bm{K_i}$ denote the $i$-th column of this matrix, we find that $\Gamma$ is defined by nonnegativity constraints and the conic constraint
$$\bm{K}-\sum_{i=1}^n \gamma_i \bm{K_i}\bm{K_i}^\top\succeq 0.$$

Finally, a formulation analogous to the one in Theorem~\ref{theo:conicPrimal} is
\begin{subequations}
	\begin{align*}
		\min_{\bm\alpha,\bm z,\bm A}\;&\langle \bm{K},\bm{A}\rangle+\lambda\sum_{i=1}^nz_i\\
		\text{s.t. }&\langle\bm{K_i}\bm{K_i}^\top, \bm{A}\rangle-2\bm{K_i^\top \alpha}+1\geq 
		\frac{(1-\bm{K_i^\top \alpha})_+^2}{z_i}+\frac{(1-\bm{K_i^\top \alpha})_-^2}{1-z_i}
		&\forall i\in [n]\\
		&\begin{pmatrix}1&\bm{\alpha^\top}\\\bm{\alpha}&\bm{A}\end{pmatrix}\succeq 0,\;\bm{\alpha}\in \R^n,\; \bm{A}\in \R^{n\times n}\\
		&\bm{z}\in [0,1]^n.
	\end{align*}
\end{subequations}


\section{Additional computational results with separable synthetic data and label noise}\label{sec:compAdditional}
We also tested the proposed methods in problems where the data is originally completely separable, but the labels of some points are flipped, obfucating the learning process. Specifically, let $0\leq \tau\leq 0.5$ be a noise parameter, and let $n,p\in \Z_+$ be dimension parameters. To create the instances we first generate a ``true" hyperplane $\bm{\bar w}\in \R^{p+1}$ separating the data, where each entry of $\bm{\bar w}$ is generated independently from a uniform distribution in $[-1,1]$.  Then we generate $n$ points $x_i\in \R^{p+1}$ where $(x_i)_1=1$ (to account for the intercept) and each $(x_i)_j$ is generated independently from a uniform distribution in $[-1,1]$. If $\bm{\bar w^\top x_i}\geq 0$ we set label $y_i=1$ with probability $1-\tau$ and $y_i=-1$ otherwise; and if $\bm{\bar w^\top x_i}< 0$ we set label $y_i=-1$ with probability $1-\tau$ and $y_i=1$ otherwise. Then we generate training, validation and testing sets and perform crossvalidation with the hinge and conic loss, as discussed in \S\ref{sec:methods}. The results are summarized in Table~\ref{tab:misclassification1Synt}.

\begin{table}[!h]
	\renewcommand{\arraystretch}{1.2}
	\caption{\small Out-of-sample misclassification rate in instances of type 1, as a function of the noise parameter $\tau$.} 
	\label{tab:misclassification1Synt}
	\setlength{\tabcolsep}{2pt}
	\centering
	\begin{tabular}{ c | c |c |c }
		\hline
		\textbf{method}&$\bm{\tau=0.1}$&$\bm{\tau=0.2}$&$\bm{\tau=0.3}$\\
		\hline
		hinge&$2.7\%\pm 2.5\%$ &$4.9\%\pm 5.4\%$&$7.1\%\pm 7.2\%$\\
		conic&$3.1\%\pm 3.1\%$&$4.5\%\pm 4.5\%$&$6.4\%\pm 6.6\%$\\\hline
	\end{tabular}
\end{table}

The results are consistent with computational results presented elsewhere in the paper: the hinge loss exhibits slightly better performance in noise-free regimes, but is inferior in the presence of outliers.

\section{Cross-validation for SVMs with the conic loss}\label{sec:crossvalidation}
In practice, we would like to solve \eqref{eq:primal} for several values of $\lambda$ and choose the best hyperparameter via cross-validation. However, the magnitude of the best parameters is often unknown and data-dependent. When doing cross-validation, instead of directly using \eqref{eq:primal}, we remove the penalty term $\lambda\sum_{i=1}^n z_i$ from the objective and instead add the constraint $$\sum_{i=1}^nz_i\leq \kappa n,$$ where $\kappa\in [0,1]$ can be loosely interpreted as the proportion of points that can be misclassified. Since the problems are convex, the two problems are equivalent in the sense that for every $\kappa$ there exists a $\lambda$ that delivers the same solution, and vice versa. However, by this transformation, when performing cross-validation, we simply select values of $\kappa$ uniformly in the interval $[0,0.5]$, matching the prior that there should never be more than 50\% misclassified points and thus testing only relevant values of the hyperparameters. We stress that the reason this approach works for the proposed conic formulation (but not for the Hinge loss) is that the formulation is a stronger formulation of the exact 0-1 loss problem \eqref{eq:BigMPenalty} (but with a cardinality constraint), thus hyperparameters such as $\kappa$ retain approximately their conceptual meanings. 

\end{document}